\documentclass[letterpaper,10pt,journal,twoside]{IEEEtran}
\usepackage{ieee_robotics}
\newcommand{\method}{B\(^{*}\)\xspace}
\newcommand{\transpose}{\mathsf{T}}

\acrodef{dof}[DoF]{degree of freedom}
\acrodefplural{dof}[DoFs]{degrees of freedom}
\acrodef{ompl}[OMPL]{Open Motion Planning Library}
\acrodef{ik}[IK]{inverse kinematics}
\acrodef{sqp}[SQP]{sequential quadratic program}
\acrodef{slp}[SLP]{sequential linear program}
\acrodef{lp}[LP]{linear program}
\acrodef{rm}[RM]{reachability map}
\acrodef{ate}[ATE]{absolute trajectory error}

\usepackage{orcidlink}
\usepackage[detect-weight,mode=text]{siunitx}
\crefname{ineq}{Ineq.}{Ineqs.}

\newcommand{\SuppVideoSI}{https://vimeo.com/1050685543}
\newcommand{\SuppVideoSII}{https://vimeo.com/1096935900}
\newcommand{\SuppVideoSIII}{https://vimeo.com/1096939247}

\title{\method: Efficient and Optimal Base Placement\\for Fixed-Base Manipulators}

\author{Zihang Zhao\orcidlink{0000-0003-3215-7152}, Leiyao Cui\orcidlink{0009-0009-4925-6983}, Sirui Xie\orcidlink{0009-0003-9379-2122}, Saiyao Zhang\orcidlink{0009-0007-2130-9110}, Zhi Han\orcidlink{0000-0002-8039-6679}, Lecheng Ruan\orcidlink{0000-0001-5061-3575}, and Yixin Zhu\orcidlink{0000-0001-7024-1545}

\vspace{-12pt}

\thanks{Manuscript received: April 16, 2025; Revised: June 22, 2025. Accepted August 22, 2025.}
\thanks{This paper was recommended for publication by Editor Chao-Bo Yan upon evaluation of the Associate Editor and Reviewers’ comments. This work is supported in part by the National Science and Technology Major Project (2022ZD0114900), the National Natural Science Foundation of China (62376009), the Beijing Nova Program, the State Key Lab of General AI at Peking University, the PKU-BingJi Joint Laboratory for Artificial Intelligence, and the National Comprehensive Experimental Base for Governance of Intelligent Society, Wuhan East Lake High-Tech Development Zone. (Z. Zhao, L. Cui, and S. Xie contributed equally to this work. Corresponding author: Yixin Zhu.)
}%
\thanks{Zihang Zhao is with the Institute for Artificial Intelligence, Peking University, Beijing 100871, China; School of Psychological and Cognitive Sciences, Peking University, Beijing 100871, China; Beijing Key Laboratory of Behavior and Mental Health, Peking University, Beijing 100871, China; and the LeapZenith AI Research, Shanghai 201707, China}%
\thanks{Leiyao Cui and Saiyao Zhang are with the University of Chinese Academy of Sciences, Beijing 100049, China, and interned at the School of Psychological and Cognitive Sciences, Peking University, Beijing 100871, China.}%
\thanks{Sirui Xie and Yixin Zhu are with the Institute for Artificial Intelligence, Peking University, Beijing 100871, China; School of Psychological and Cognitive Sciences, Peking University, Beijing 100871, China; and Beijing Key Laboratory of Behavior and Mental Health, Peking University, Beijing 100871, China (email: \texttt{yixin.zhu@pku.edu.cn}).}%
\thanks{Zhi Han is with the University of Chinese Academy of Sciences, Beijing 100049, China.}%
\thanks{Lecheng Ruan is with the College of Engineering, Peking University, Beijing 100871, China.}%
\thanks{Data is available online at \url{https://bstar-planning.github.io}.}
\thanks{Digital Object Identifier (DOI): see top of this page.}
}

\begin{document}
\maketitle
\begin{abstract}
Proper base placement is crucial for task execution feasibility and performance of fixed-base manipulators, the dominant solution in robotic automation.
Current methods rely on pre-computed kinematics databases generated through sampling to search for solutions. However, they face an inherent trade-off between solution optimality and computational efficiency when determining sampling resolution---a challenge that intensifies when considering long-horizon trajectories, self-collision avoidance, and task-specific requirements.
To address these limitations, we present \method, a novel optimization framework for determining the optimal base placement that unifies these multiple objectives without relying on pre-computed databases.
\method addresses this inherently non-convex problem via a two-layer hierarchical approach: The outer layer systematically manages terminal constraints through progressively tightening them, particularly the base mobility constraint, enabling feasible initialization and broad solution space exploration. Concurrently, the inner layer addresses the non-convexities of each outer-layer subproblem by sequential local linearization, effectively transforming the original problem into a tractable \acf{slp}.
Comprehensive evaluations across multiple robot platforms and task complexities demonstrate the effectiveness of \method: it achieves solution optimality five orders of magnitude better than sampling-based approaches while maintaining perfect success rates, all with reduced computational overhead.
Operating directly in configuration space, \method not only solves the base placement problem but also enables simultaneous path planning with customizable optimization criteria, making it a versatile framework for various robotic motion planning challenges.
\method serves as a crucial initialization tool for robotic applications, bridging the gap between theoretical motion planning and practical deployment where feasible trajectory existence is fundamental. 
\end{abstract}

\begin{figure}[b!]
    \centering
    \includegraphics[width=\linewidth]{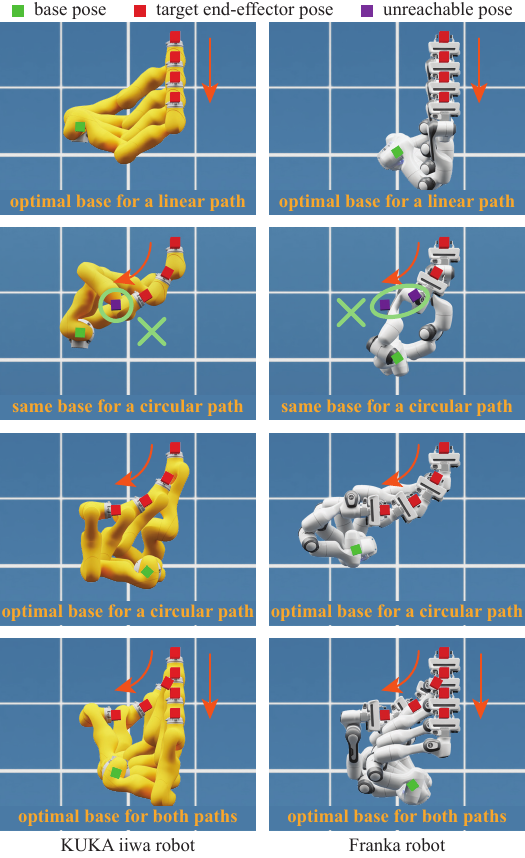}
    \caption{\textbf{Varied optimal base placement across robotic platforms and task requirements.} To demonstrate this, we present several examples using two widely adopted manipulators---the KUKA iiwa (left) and Franka (right) robots. Our examples encompass four key scenarios: (i) optimal base placement for a linear path, (ii) a circular path using the base placement optimized for the linear path (showing unreachable poses), (iii) optimal base placement for the circular path, and (iv) optimal base placement that accommodates both paths. This comparison illustrates the necessity of an efficient method to generate the optimal base placement for each scenario.}
    \label{fig:teaser}
\end{figure}


\section{Introduction}

Fixed-base manipulators continue to dominate robotic automation due to their stability, precision, and payload capacity~\cite{ferreira202221st,zhao2025embedding}, despite emerging advances in mobile and humanoid platforms~\cite{arm2023scientific,dafarra2024icub3}. Supported by well-established commercial offerings, they remain central to both industrial applications and research advances. This progress is further accelerated by sophisticated algorithms, from global motion planning~\cite{dijkstra2022note,hart1968formal,sucan2012open,zucker2013chomp,schulman2014motion} to local motion generation~\cite{karayiannidis2016adaptive,o2024open,zhao2024tac}, which autonomously generate joint trajectories for diverse tasks.

The effectiveness of these algorithms, however, depends critically on well-defined Cartesian trajectories within the manipulator's workspace~\cite{buss2005selectively}. For fixed-base manipulators, this workspace is determined by the base placement, making proper base placement crucial to successful trajectory execution. The significance of base positioning extends beyond basic reachability considerations---it directly influences the manipulator's ability to maintain continuous joint configurations between via points, avoid self-collision, and optimize performance throughout complete motion sequences.

This placement problem presents complex challenges due to the highly non-convex nature of robot workspaces, characterized by discontinuities, singularities, and regions of varying dexterity~\cite{zacharias2007capturing,shetty2024tensor}. As illustrated in \cref{fig:teaser}, optimal base placement is highly sensitive to both the robot's kinematic structure and specific task requirements---a configuration that minimizes trajectory length for one task may severely limit the robot's capabilities for another.

Current base placement methods face a fundamental limitation: they rely on pre-computed kinematic databases generated through discretized sampling of the robot's workspace. These databases---storing either direct mappings between end-effector and base positions or collections of \ac{ik} solutions (see \cref{sec:related_work})---create an unavoidable trade-off. High-resolution sampling produces better solutions but severely impacts computational efficiency; low-resolution sampling offers speed but compromises solution quality and feasibility. This limitation is intrinsic to the sampling-based paradigm.

Furthermore, as task complexity grows---such as requiring more via points in a path, incorporating self-collision checks, or meeting additional task-specific constraints---the scalability of these methods is further compromised. This results in a significant challenge: balancing precision and computational speed becomes increasingly difficult regardless of search algorithm sophistication. These limitations motivate the development of a new framework that can address these shortcomings and provide a more efficient and scalable solution for base placement optimization.

We present \method, an optimization framework that overcomes these limitations by unifying path-wide feasibility, self-collision avoidance, and task-specific requirements into a continuous optimization formulation. To tackle this highly complex optimization, \method employs two-layer hierarchical strategies:
\begin{itemize}[leftmargin=*,noitemsep,nolistsep]
    \item Outer layer with progressive constraint tightening: The terminal optimization problem is temporarily reformulated into a more tractable form through systematic constraint relaxation, particularly the base mobility constraint. Initially, the fixed-base constraint is relaxed by introducing three additional \acp{dof} to model the base, akin to a mobile manipulator. This relaxation ensures feasible initialization through guaranteed \ac{ik} solutions while enabling the exploration of a broader solution space. Constraints are then gradually enforced, using the explored solutions as effective initial conditions for subsequent optimization.    
    \item Inner layer with sequential linearization: To address the inherent non-convexities within each sub-problem generated by the outer layer, an iterative process of local linearization is employed. This approach approximates the non-convex problem by constructing a sequence of locally linear sub-problems that are computationally tractable. Combined with the progressive constraint-tightening strategy of the outer layer, this hierarchical framework effectively transforms the originally highly non-convex optimization problem into a tractable \acf{slp}.
\end{itemize}

We quantitatively validate \method through comprehensive testing in four manipulator types and \num{2400} randomly generated paths of varying complexity levels. Operating directly in configuration space without relying on pre-computed databases, \method achieves: (i) perfect success rates across all test cases, even for complex trajectories with multiple via-points and self-collision constraints; (ii) solution optimality five orders of magnitude better than sampling-based methods, particularly in challenging scenarios requiring precise positioning; and (iii) reduced computational overhead despite higher precision.

These results establish \method as a versatile framework for optimal base placement, effectively bridging the gap between theoretical motion planning and practical deployment. By operating directly in configuration space, our approach not only solves the base placement problem but also enables simultaneous path planning with customizable optimization criteria. This capability opens new possibilities for unified trajectory and base optimization in robotic manipulation, particularly in applications requiring high precision and efficiency.

Our contributions are twofold:
\begin{itemize}[leftmargin=*,noitemsep,nolistsep]
\item A novel configuration-space optimization framework that determines optimal base placement without precomputed databases, ensuring solution optimality while maintaining high success rates and computational efficiency.

\item Comprehensive evaluation across multiple manipulators (6-\acs{dof} to 7-\acs{dof}) and varying path complexities, demonstrating the framework's versatility and robustness in diverse scenarios.
\end{itemize}

The remainder of this paper is organized as follows: \cref{sec:related_work} examines current base placement methods and their limitations in detail. \cref{sec:method} presents the theoretical framework of \method, including our novel optimization approach. \cref{sec:simulation} provides comprehensive experimental validation across various robotic platforms and task complexities. \cref{sec:discussion} analyzes key performance characteristics and discusses future research directions. \cref{sec:conclusion} summarizes our findings and contributions.

\section{Related Work}\label{sec:related_work}

A proper base placement is fundamental for fixed-base manipulators to execute tasks successfully. Research in this field has evolved through several stages of increasing complexity, starting with simple reachability verification and progressing to sophisticated trajectory-aware optimization.

Initial studies concentrated on single-point position reachability~\cite{vahrenkamp2013robot}, leveraging the concept of \acfp{rm}. These maps were precomputed by systematically sampling the kinematic relationships between the base and end-effector, expressed through homogeneous transformation matrices~\cite{zacharias2007capturing,birr2022oriented,rudorfer2024rm4d}. While this foundational approach proved effective for basic positioning, researchers recognized the need for optimization criteria, leading to the integration of manipulability metrics for base placement selection~\cite{burget2015stance,reister2022combining}. The limitations of static reachability maps in dynamic environments prompted the development of adaptive \acp{rm}~\cite{yang2016idrm}. Building upon these advances, researchers expanded the framework to address more sophisticated requirements, incorporating base orientation considerations~\cite{dong2015orientation,birr2022oriented} and multi-point reachability analysis~\cite{paus2017combined,xu2020planning}. Recent advances in learning-based methodologies have enabled neural networks to function as efficient approximators for complex kinematic mappings, substantially enhancing computational performance for single-point reachability analysis~\cite{sandakalum2022inv}.

Despite significant advances in reachability mapping, a fundamental limitation remained: while these methods could verify point-wise reachability, they could not guarantee the feasibility of continuous trajectories between reachable points. This limitation became particularly apparent in practical applications where two points might be individually reachable yet connected by no viable path due to kinematic constraints or environmental obstacles. This crucial insight led to a paradigm shift in representing kinematic relationships, replacing the traditional homogeneous transformation matrix approach with an \ac{ik}-based formulation. This transition enabled a more comprehensive evaluation of both point reachability and path feasibility within the robot's configuration space~\cite{osswald2017efficient,weingartshofer2021optimal}.

However, even \ac{ik}-based approaches remain constrained by their reliance on pre-computed sampling-based databases and search algorithms for optimal base placement. This dependence on discrete sampling creates an inherent trade-off: higher sampling resolution provides better precision but demands greater computational resources and memory capacity, while lower resolution sacrifices precision for efficiency.

In contrast to existing approaches, this paper introduces \method, a novel optimization-based framework for determining optimal base placement. Unlike traditional methods that depend on pre-computed databases, \method operates directly within the robot's configuration space. Moreover, compared to learning-based methods, it offers broader applicability to manipulators with arbitrary configurations. Through its unified optimization formulation, \method achieves significant improvements in computational efficiency while maintaining high success rates and solution optimality across diverse task scenarios.

\section{The \texorpdfstring{\method}{} Method}\label{sec:method}

We present \method, our optimization framework for fixed-base manipulator placement. We first formalize the mathematical foundations of optimal base placement (\cref{sec:method-formulation}), then introduce a two-layer optimization strategy that decomposes this non-convex problem into tractable sub-problems (\cref{sec:method-two_layer}).

\subsection{Problem Formulation}\label{sec:method-formulation}

We formulate the base placement problem for fixed-base manipulators, intended to be mounted on a flat surface (whether horizontal or slanted), as an optimization problem. For a robotic manipulator mounted on a fixed base, we seek to compute the optimal base placement \(\boldsymbol{q}^b=[x^{b},y^{b},\theta^{b}]^{\transpose}\in \textit{SE}(2)\) (\(\textit{SE}(k)\) denotes special Euclidean group in \(k\) dimensions) that enables successful task execution. The manipulation task comprises an ordered sequence of desired end-effector poses in \(\textit{SE}(3)\):
\begin{equation}
    \boldsymbol{x}_{1:t} = [\boldsymbol{x}_{1},\boldsymbol{x}_{2},\cdots,\boldsymbol{x}_{t}]\in\mathbb{R}^{t\times 6},
\end{equation}
where each \(\boldsymbol{x}_i \in \mathbb{R}^6\) encodes position and orientation parameters at time step \(i\). These task poses can be provided directly by a human operator or generated via interpolation methods for smooth and continuous motion between successive steps.

For any candidate base placement \(\boldsymbol{q}^b\), the manipulator must achieve feasible joint configurations \(\boldsymbol{q}^{m}_{1:t}=[\boldsymbol{q}_{1},\boldsymbol{q}_{2},\cdots,\boldsymbol{q}_{t}]\in\mathbb{R}^{t\times n}\) to reach each target pose \(\boldsymbol{x}_i\), where \(n\) is \ac{dof} of the manipulator. The optimization problem requires finding both an optimal base placement and a sequence of valid joint configurations that satisfy the manipulator's kinematic constraints. This leads to the following formulation:
\begin{equation}
    \begin{aligned}
        \text{minimize} \quad{}& f(\boldsymbol{q}^b,\boldsymbol{q}^{m}_{1:t}) \\
        \text{subject to} \quad{}& g_i(\boldsymbol{q}^b,\boldsymbol{q}^{m}_{1:t}) = 0, \quad i = 1, 2, \ldots, n_{\text{eq}}\\
         &h_i(\boldsymbol{q}^b,\boldsymbol{q}^{m}_{1:t}) \leq 0, \quad i = 1, 2, \ldots, n_{\text{ineq}}
    \end{aligned}
\end{equation}
where \(f\) is a scalar objective function, and \(g_i\) and \(h_i\) represent equality and inequality constraints, respectively. We detail these objectives and constraints below.

\paragraph*{Objective Function}

The primary requirement in base placement optimization is satisfying reachability constraints while maintaining flexibility in objective function formulation. We present two common formulations. The simplest is the feasibility-only approach:
\begin{equation}
    f(\boldsymbol{q}^b,\boldsymbol{q}^{m}_{1:t}) = 1,
\end{equation}
which transforms the optimization into a feasibility test, seeking any valid solution that satisfies all constraints.

For tasks requiring efficient execution, we can further minimize path length:
\begin{equation}    f(\boldsymbol{q}^b,\boldsymbol{q}^{m}_{1:t}) = \sum_{i=1}^{t-1} \|\boldsymbol{q}^{m}_{i+1}-\boldsymbol{q}^{m}_{i}\|_1.
    \label{eq:minimal_length}
\end{equation}

We utilize non-differentiable \(\ell_1\) penalties, although smooth \(\ell_2\) penalties produce comparable results~\cite{schulman2014motion}. The inclusion of the \(\ell_1\) term elegantly reformulates the problem into an equivalent linear programming task with linear constraints. By introducing auxiliary slack variables \(z_i\) with constraints \(z_i \geq x_i\), \(z_i \geq -x_i\) and \(z_i \geq 0\), each absolute value term \(|x_i|\) is transformed into a linear program equivalent. This choice also ensures compatibility with a wide range of optimization solvers~\cite{diamond2016cvxpy}.

\paragraph*{Equality Constraints}

The primary equality constraints ensure the end-effector reaches all targeted poses. Let \(\psi(\boldsymbol{q}^b,\boldsymbol{q}^m_i):\textit{SE}(2)\times\mathbb{R}^{n}\rightarrow\mathbb{R}^{6}\) denote the forward kinematics mapping from the base and joint configuration to end-effector pose. We enforce exact pose achievement through:
\begin{equation}
    g_1: \psi(\boldsymbol{q}^b, \boldsymbol{q}^m_{i}) = \boldsymbol{x}_i,\quad i = 1, 2, \ldots, t.
    \label{eq:EE_constraint}
\end{equation}

\paragraph*{Inequality Constraints} 

In addition to achieving each pose, the manipulator must respect several physical and safety constraints. First, base placement \(\boldsymbol{q}^b \in \textit{SE}(2)\) must remain within allowable position and orientation ranges determined by working conditions. Let \({\boldsymbol{q}^b}^{m}\) and \({\boldsymbol{q}^b}^{M}\) denote the lower and upper bounds, respectively:
\begin{equation}
    h_1: {\boldsymbol{q}^b}^{m} \preccurlyeq \boldsymbol{q}^b \preccurlyeq {\boldsymbol{q}^b}^{M}, 
    \label{eq:base_limit}
\end{equation}
where \(\preccurlyeq\) represents a element-wise ``less than or equal to'' relationship between vectors.

Next, each joint configuration \(\boldsymbol{q}^m_i\) must remain within its feasible range, bounded by mechanical joint limits. With lower and upper bounds denoted as \({\boldsymbol{q}^m}^m\) and \({\boldsymbol{q}^m}^M\), we impose:
\begin{equation}
    h_2: {\boldsymbol{q}^m}^{m} \preccurlyeq \boldsymbol{q}^m_i \preccurlyeq {\boldsymbol{q}^m}^{M}, \quad i = 1, 2, \ldots, t.
\end{equation}

Finally, collision-free motion is enforced through:
\begin{equation}
    h_3: \text{sd}(\boldsymbol{q}^b,\boldsymbol{q}^{m}_{i})\geq 0,\quad i = 1, 2, \ldots, t,
\end{equation}
where the signed distance \(\text{sd}(\cdot)\) as defined in \cite{ericson2004real} represents the minimal translation distance required to alter the spatial relationship between objects. This formulation addresses self-collision avoidance and can be extended to obstacle avoidance by incorporating obstacle geometry information.

\subsection{Two-Layer Optimization of \texorpdfstring{\method}{}}\label{sec:method-two_layer}

The optimization problem presented above is highly non-convex and challenging to solve directly. We address this complexity through a hierarchical two-layer optimization structure. Specifically, the inner layer manages local non-convexity through iterative convex approximations of the original problem. The outer layer employs a novel approach by initially treating the base as three additional \acp{dof}, analogous to a mobile manipulator. This formulation ensures solution feasibility while enabling comprehensive exploration of the solution space. By progressively increasing penalties on base movement, the algorithm converges to a fixed base configuration, effectively balancing thorough solution space exploration with fixed-base constraint satisfaction.

\paragraph*{Outer Layer---Base Relaxation}

The complexity of optimization stems largely from finding an initially feasible solution under fixed base constraints. We address this through a strategic relaxation approach.

We first solve a relaxed problem by treating the fixed base as mobile, transforming the timestep-invariant base configuration \(\boldsymbol{q}^b\) into a time-varying sequence \(\boldsymbol{q}^b_{1:t}\), effectively introducing three additional \ac{dof} per timestep. To ensure convergence to a fixed base configuration, we introduce a base movement penalty with an iteration-dependent coefficient \(\mu(j)\) that increases with outer loop iteration \(j\).

The modified objective function \(f'(\boldsymbol{q}^b_{1:t},\boldsymbol{q}^{m}_{1:t};\mu(j))\) becomes:
\begin{equation}
    f'(\boldsymbol{q}^b_{1:t},\boldsymbol{q}^{m}_{1:t};\mu(j)) = f(\boldsymbol{q}^b_{1:t},\boldsymbol{q}^{m}_{1:t}) + \mu(j)\sum_{i=1}^{t}\|\boldsymbol{q}^{b}_{i}-\bar{\boldsymbol{q}}^{b}\|_1,
\end{equation}
where \(\bar{\boldsymbol{q}}^{b}\) is the arithmetic mean of base poses across all steps. We employ \(\ell_1\) penalties over \(\ell_2\) as they more effectively drive the second term to zero with matching coefficients \(\mu(j)\), ensuring convergence to a stable configuration. The coefficient \(\mu(j)\) progressively increases across outer loop iterations, gradually enforcing the fixed base constraint.

All constraints involving the fixed base \(\boldsymbol{q}^b\) must be reformulated for the relaxed problem. Each instance of \(\boldsymbol{q}^b\) in the original constraints is replaced with its time-indexed counterpart \(\boldsymbol{q}^b_i\) at the corresponding timestep \(i\). For example, the base limits constraint becomes:
\begin{equation}
    h_1: {\boldsymbol{q}^b}^{m} \preccurlyeq \boldsymbol{q}^b_i \preccurlyeq {\boldsymbol{q}^b}^{M}, \quad i = 1, 2, \ldots, t.
\end{equation}
These constraints are subsequently transformed into penalty terms to accommodate potentially infeasible initial conditions, ensuring the algorithm can start from arbitrary initial states while maintaining numerical stability.

\paragraph*{Inner Layer---Linearization}

Even after relaxation, the problem's inherent non-convexity poses significant challenges for optimization. Drawing inspiration from sequential programming~\cite{nocedal1999numerical}, we address this through iterative convex approximations within trust regions.

Given a non-convex function \(\phi(x)\), we construct its convex approximation \(\phi_c(x)\) through first-order Taylor expansion around the current point \(x_0\):
\begin{equation}
    \phi_c(x) = \phi(x_0) + \dot{\phi}(x_0)(x-x_0),
\end{equation}
where \(\dot{\phi}(x_0)\) represents the first-order derivative at \(x_0\). The trust region size adapts based on approximation quality---expanding when the approximation performs well and contracting when it poorly represents the original function~\cite{nocedal1999numerical}.

This two-layer framework combines base relaxation with local linear approximation to transform the non-convex optimization into a series of tractable \acp{lp}. The outer iterations broadly explore the solution space for suitable initialization, while inner iterations efficiently traverse the current non-convex space to determine a high-quality solution specifying fixed base placement and joint configurations.

\begin{figure*}[t!]
    \centering
    \includegraphics[width=\linewidth]{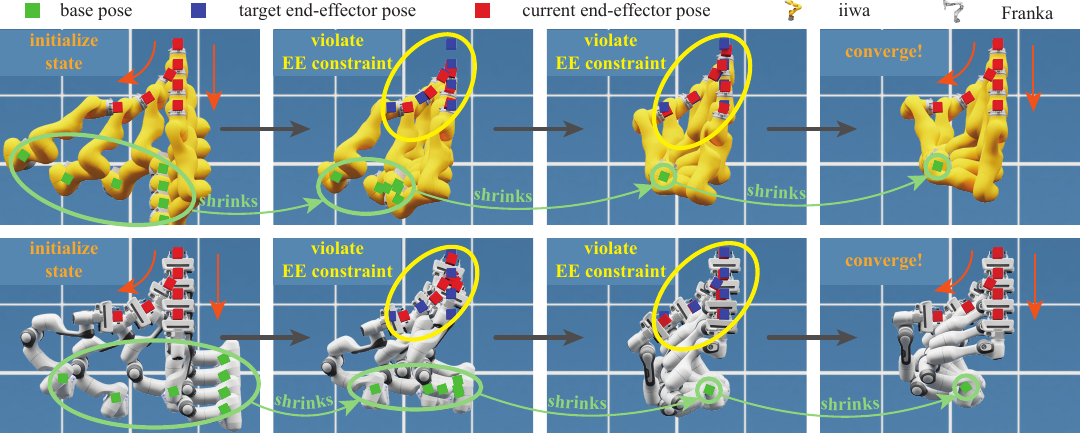}
    \caption{\textbf{Visualization of \method's optimization process over multiple iterations.} Each row shows a different test scenario, demonstrating the method on both the KUKA iiwa and Franka robots performing compound trajectories. Starting from an initial configuration with relaxed base constraints (leftmost column), the optimization progressively constrains the base position through increasing penalty coefficients (middle columns), until converging to a single fixed base placement (rightmost column). During intermediate steps, some constraints, such as the end-effector (EE) constraint (\cref{eq:EE_constraint}), may temporarily violate the target states as the algorithm balances between base convergence and task constraints. More examples and full optimization sequences are available in the \href{\SuppVideoSI}{Supplementary Video S1}.}
    \label{fig:results-progress}
\end{figure*}

\section{Evaluations}\label{sec:simulation}

We evaluate \method through comprehensive simulation studies assessing its effectiveness in determining optimal fixed-base manipulator poses across diverse tasks. Our evaluation comprises two components: First, \cref{sec:simulation-qualitative_results} provides qualitative insights into \method's operational principles through representative examples that illustrate its behavior across different scenarios. Second, \cref{sec:simulation-quantitative_results} presents detailed quantitative comparisons against baseline methods using extensive randomly generated test cases of varying complexity.

\subsection{Implementation Details}\label{sec:simulation-computation}

\method optimizes for minimal-length task completion paths as formalized in \cref{eq:minimal_length}. The process initializes using an \ac{ik} solution incorporating 3 additional base \acp{dof}. For the first point \(\boldsymbol{x}_1\), random initialization ensures comprehensive solution space exploration. Subsequent points are initialized using the preceding point's \ac{ik} solution to promote trajectory continuity.

The optimization problem is implemented using the COPT solver~\cite{ge2022cardinal}. Given the established correlation between initialization quality and sequential programming performance~\cite{schulman2014motion}, \method employs up to 10 retry attempts with different initializations when needed.

\subsection{Qualitative Results}\label{sec:simulation-qualitative_results}

We evaluated \method through experiments on a representative trajectory type, incorporating ground collision avoidance constraints. The type comprises complex compound paths that combine linear and circular segments, requiring 64 uniformly spaced end-effector poses. To demonstrate generalizability, we tested \method on two industrial manipulators with distinct kinematic structures and workspace characteristics: the Franka Emika Panda and KUKA LBR iiwa.

\cref{fig:results-progress} captures the optimization progression through four key iterations for each task category. To maintain visual clarity, we show selected configurations---7 poses for compound paths. Starting from the leftmost column with initial \ac{ik} solutions using relaxed base poses (\cref{sec:simulation-computation}), the optimization proceeds through intermediate stages where increasing penalty coefficients drive convergence toward a fixed base. During this process, the robot may temporarily deviate from target end-effector poses due to our soft constraint formulation. The rightmost column shows the final solution: a single, collision-free base placement that enables the robot to reach all desired end-effector poses successfully.

\vspace{-12pt}

\subsection{Quantitative Results}\label{sec:simulation-quantitative_results}

We conducted extensive quantitative evaluations comparing \method against baseline approaches across varying task complexities and robot platforms. Our evaluation framework consists of four key components: (i) a systematic test data generation process covering diverse workspace configurations, (ii) implementation of established baseline methods using pre-computed \ac{ik} databases, (iii) comprehensive performance metrics measuring success rate, solution optimality, and computational efficiency, and (iv) controlled testing environment for fair comparison.

\paragraph*{Large-Scale Test Data Generation}

To ensure rigorous evaluation, we developed a systematic approach for generating test datasets that comprehensively span each manipulator's workspace across varying complexity levels. Our data generation process consists of three key components.

The first component focuses on initial configuration sampling. We generate the start pose \(\boldsymbol{x}_1\) by uniformly sampling the joint space to achieve comprehensive coverage:
\begin{equation}
    \boldsymbol{x}_1 = \psi(\boldsymbol{q}^b,\boldsymbol{q}_1^m),\quad \boldsymbol{q}_1^m\sim\mathcal{U}({\boldsymbol{q}^{m}}^{m},{\boldsymbol{q}^{m}}^{M}).
\end{equation}
\(\boldsymbol{q}^{b}\) merely accounts for a planar transformation in \(\textit{SE}(2)\) without affecting workspace characteristics. Therefore, we set \(\boldsymbol{q}^{b}\) to zero vector without losing generality.

The second component generates workspace-valid pose sequences. To maintain kinematic feasibility, we construct subsequent poses through incremental perturbations in joint space. Each pose builds upon its predecessor through small displacements \(\boldsymbol{\xi}_j\):
\begin{equation}
    \boldsymbol{x}_k = \psi(\boldsymbol{q}^b,\boldsymbol{q}_1^m + \sum_{j=1}^{k-1}\boldsymbol{\xi}_j),\quad \boldsymbol{\xi}_j\sim\mathcal{N}(\boldsymbol{\mu},\boldsymbol{\Sigma}),
\end{equation}
where \(\boldsymbol{\mu}\in\mathbb{R}^{n}\) has all entries equal to \SI{0.01}{\radian} and \(\boldsymbol{\Sigma}=\text{diag}(0.005^2)\in\mathbb{R}^{n\times n}\). These parameters were empirically selected for effective workspace exploration. We validate trajectory feasibility using NVIDIA Isaac Sim's Luna module.

The third component implements multi-level complexity scaling. For systematic evaluation across different scales, we generate pose sequences of varying lengths based on complexity level \(l\):
\begin{equation}
    t = 2^l, \quad l \in \mathbb{N}^+.
    \label{eq:complexity_level}
\end{equation}

We applied this framework to four widely-used manipulators: Franka Emika Panda, Kinova Gen3, KUKA LBR iiwa, and Universal Robots UR10, covering both 6-\ac{dof} and 7-\ac{dof} manipulators. For each robot-complexity combination (\(l=1,\ldots,6\)), we generated 100 distinct sequences, yielding \num{2400} comprehensive test cases.

\begin{figure}[t!]
    \centering
    \includegraphics[width=\linewidth]{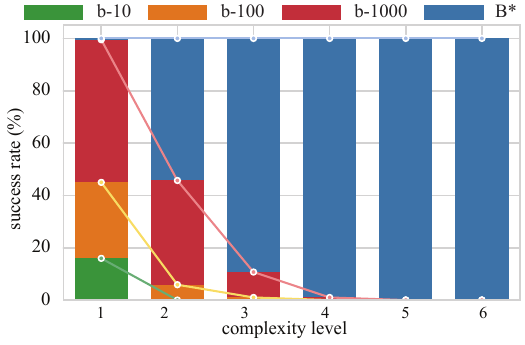}
    \caption{\textbf{Success rate comparison between baseline methods and \method.} While baseline methods with more \acs{ik} samples show higher success rates at low complexity levels, their performance deteriorates sharply as the complexity level \(l\) increases. With just 10 samples (b-10), the baseline struggles even at level 1, while 1000 samples (b-1000) maintain moderate success through level 2 before failing. In contrast, \method maintains 100\,\% success rate across all complexity levels \(l\) (see \cref{eq:complexity_level}).}
    \label{fig:results-success_rate}
\end{figure}

\paragraph*{Baseline Methods}

We evaluated \method against search-based approaches utilizing pre-computed \ac{ik} solution databases. These baseline methods treat the base position's three planar \ac{dof} as additional joints in the \ac{ik} formulation. For each target point in a sequence, we employ CuRobo~\cite{sundaralingam2023curobo} to generate \(\gamma\) collision-free \ac{ik} solutions comprising both manipulator joint configurations and base positions. Using this \ac{ik} dataset, the baseline methods search for feasible joint paths allowing point-to-point movement while maintaining the base position within a confined space. We implement this search using a breadth-first algorithm to ensure global optimality for fair comparison with the optimization-based method \method in solution optimality. To investigate the impact of solution sampling density, we evaluated three baseline configurations: b-10, b-100, and b-1000, corresponding to \(\gamma \in \{10, 100, 1000\}\) \ac{ik} solutions per target point.

\begin{figure}[t!]
    \centering
    \includegraphics[width=\linewidth]{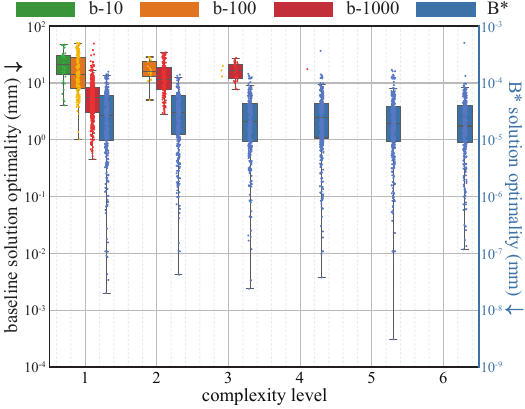}
    \caption{\textbf{Solution optimality (\acs{ate}) comparison between baseline methods (left axis) and \method (right axis) across complexity levels.} Baseline methods, even with increased \acs{ik} samples, achieve only millimeter-level precision due to sampling resolution limits. \method achieves five orders of magnitude better precision by operating in continuous configuration space. Box plots show error distributions: boxes indicate interquartile range (IQR), center lines represent medians, and whiskers extend to \(1.5\times\)IQR beyond quartiles.}
    \label{fig:solution_quality}
\end{figure}

\paragraph*{Metrics}

To comprehensively evaluate performance across methods, we employ three complementary metrics that assess different aspects of solution quality and efficiency:
\begin{itemize}[leftmargin=*,noitemsep,nolistsep]
    \item \textit{Success rate:} As our primary effectiveness metric, success rate evaluates algorithmic reliability using well-defined binary criteria. For \method, success is determined by optimization convergence to a valid solution. Baseline methods must satisfy two conditions for success: maintaining base position variations within specified tolerances (\SI{0.01}{\m} for translations and \SI{0.05}{\radian} for rotation), and completing the search within a \SI{10}{\minute} time limit.
    \item \textit{Solution optimality:} To rigorously assess solution optimality, we quantify base path deviation using the \acf{ate}~\cite{sturm2012benchmark}. This metric computes the root-mean-squared error of Lie algebra components across all pairwise base placement combinations.
    \item \textit{Runtime:} Computational scalability serves as a crucial criterion for practical deployment in complex scenarios. We analyze time efficiency by measuring total computation times across varying complexity levels. For \method, this encompasses the complete runtime of all optimization attempts. For baseline methods, we include both the time required for \ac{ik} solution generation and the subsequent optimal base placement search process.
\end{itemize}

\paragraph*{Test Environment}

The implementation of all methods uses Python as the primary programming language, with all computations performed exclusively on the CPU using an AMD Ryzen 9 5950X processor supported by \SI{64}{\giga\byte} of RAM. 

\begin{figure}[t!]
    \centering
    \includegraphics[width=\linewidth]{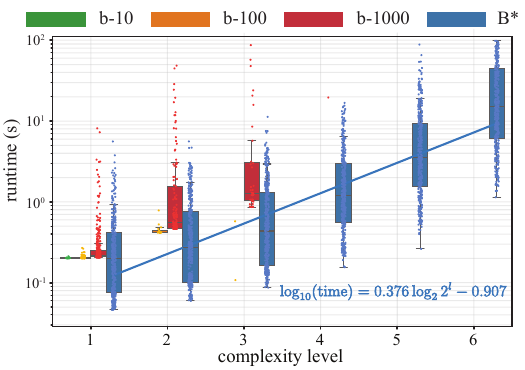}
    \caption{\textbf{Runtime comparison between baseline methods and \method across complexity levels}. While baseline methods show increased runtime with higher \ac{ik} numbers, \method demonstrates superior efficiency with linear scaling (fit shown: \(\log_{10}(\text{time}) = 0.376\log_2 2^l - 0.907\)). Box plots in the same format as \cref{fig:solution_quality} illustrate runtime distributions, with outliers above \SI{100}{\second} excluded for visualization clarity.}
    \label{fig:run_time}
\end{figure}

\begin{table}
\small
\centering
\setlength{\tabcolsep}{3pt}
\caption{\textbf{Sub-process runtime of \method in milliseconds.}}
\begin{tabular}{cccc} 
\toprule
Level & Initialization & Inner Layer & Outer Layer \\ 
\midrule
$1$ & $6.255 \pm 3.561$ & $20.516\pm3.898\text{\ \,}$ & $\text{\ \,}79.350\pm115.193$ \\
$2$ & $9.216\pm5.130$ & $25.491\pm4.870\text{\ \,}$ & $132.652\pm188.241$ \\
$3$ & $13.665\pm4.725\text{\ \,}$ & $35.049\pm6.178\text{\ \,}$ & $219.044\pm304.350$ \\
$4$ & $24.507\pm8.162\text{\ \,}$ & $60.584\pm10.931$ & $451.903\pm564.226$ \\
$5$ & $46.386\pm14.869$ & $114.614\pm21.725\text{\ \,}$ & $1032.060\pm1191.717$ \\
$6$ & $96.314\pm42.203$ & $269.816\pm65.030\text{\ \,}$ & $2838.154\pm3123.153$ \\
\bottomrule
\end{tabular}
\label{tab:run_time}
\end{table}

\begin{table}[ht!]
    \centering
    \small
    \setlength{\tabcolsep}{2.5pt}
    \caption{\textbf{\# of initialization retries across complexity levels}.}
    \begin{tabular}{lcccccc}
    \toprule
         level &$1$ &$2$ &$3$ &$4$ &$5$ &$6$ \\
    \midrule
         \# &$1.1\pm0.3$ &$1.1\pm0.3$ &$1.1\pm0.4$ &$1.3\pm0.8$ &$1.4\pm1.0$ &$2.0\pm2.0$ \\
    \bottomrule
    \end{tabular}
    \label{tab:number_of_retries}
\end{table}

\begin{figure*}[t!]
    \centering
    \begin{subfigure}[b]{.24\linewidth}
        \centering
        \includegraphics[width=\linewidth]{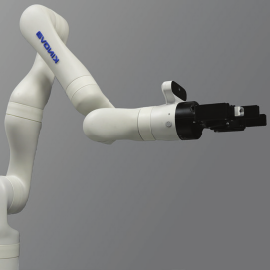}
        \caption{apparatus}
        \label{fig:real_word_exp-a}
    \end{subfigure}%
    \hfill%
    \begin{subfigure}[b]{.24\linewidth}
        \centering
        \includegraphics[width=\linewidth]{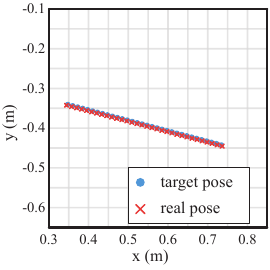}
        \caption{a linear path}
        \label{fig:real_word_exp-b}
    \end{subfigure}%
    \hfill%
    \begin{subfigure}[b]{.24\linewidth}
        \centering
        \includegraphics[width=\linewidth]{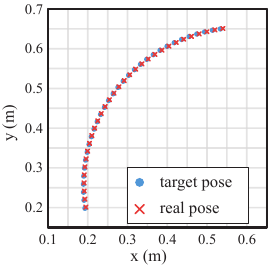}
        \caption{a circular path}
        \label{fig:real_word_exp-c}
    \end{subfigure}%
    \hfill%
    \begin{subfigure}[b]{.24\linewidth}
        \centering
        \includegraphics[width=\linewidth]{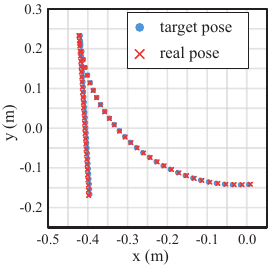}
        \caption{both paths}
        \label{fig:real_word_exp-d}
    \end{subfigure}%
    \caption{\textbf{Real-world validation.} (a) The effectiveness of \method is demonstrated on a 7-\ac{dof} Kinova robot in a real-world setting. With accurate kinematic and collision models, the joint path generated by \method is successfully executed, achieving the target pose. This is illustrated through the execution of a linear path (b), a circular path (c), and a compound path (d). The complete execution process is available at the \href{\SuppVideoSII}{Supplementary Video S2}.}
    \label{fig:real_word_exp}
\end{figure*}

\paragraph*{Results}

Our experimental evaluation reveals clear performance patterns across different methods. As shown in \cref{fig:results-success_rate,fig:solution_quality}, baseline approaches are effective for simple scenarios but degrade significantly with increasing sequence length. While using more \ac{ik} solutions improves both success rate and solution optimality, this comes at a substantial computational cost. This trade-off is evident in \cref{fig:run_time}, where higher numbers of \ac{ik} solutions result in longer runtime due to increased search complexity.

\method, in contrast, achieves optimal performance across all robots and complexity levels, maintaining consistent success rates as demonstrated in \cref{fig:results-success_rate}. Through direct configuration space optimization, \method overcomes sampling resolution limitations, achieving five orders of magnitude improvement in solution optimality (\cref{fig:solution_quality}). These substantial gains come with improved computational efficiency---\method maintains faster completion times than baseline methods while scaling linearly with scenario complexity (\cref{fig:run_time}) with detailed sub-process run time in \cref{tab:run_time}.

\section{Discussions and Future Work}\label{sec:discussion}

Our study highlights \method's effectiveness in efficiently determining optimal base placements for manipulation tasks. Traditional approaches relying on pre-computed kinematic databases show a clear dependence on the number of \ac{ik} solutions (\cref{fig:results-success_rate,fig:solution_quality}). While infinite sampling could theoretically achieve perfect results, this is impractical due to significant computational and storage overhead, especially in the breadth-first search used in baseline methods to guarantee optimality. Although alternative search methods like depth-first or heuristic search can reduce resource demands, they compromise optimality. More importantly, regardless of the search method used, these approaches fail to address the fundamental issue of sampling resolution inherent in sampling-based methods.

\method tackles these challenges with a fundamentally different approach: direct optimization in the robot’s configuration space. This method achieves a five-order-of-magnitude improvement in solution optimality over baseline methods while maintaining a perfect success rate and linear computational scalability. Beyond base placement optimization, the \method also serves as a path planner with customizable cost functions, enabling the optimization of additional criteria, such as minimal path length, as demonstrated in this study.

\method incorporates constraints such as joint limits, ground collisions, and self-collisions into the optimization. With accurate kinematic and collision models, its results are transferable to real-world applications. To validate this, we applied \method to a physical 7-DoF Kinova robot with target poses shown in \cref{fig:teaser}. For convenience and without compromising rigor, we transform the world coordinate frame to align with the base poses calculated by \method in each scenario. The generated joint configurations were directly executed on the physical robot, with end-effector poses verified through the Kinova official interface. The results, presented in \cref{fig:real_word_exp}, confirm that all configurations are executable and achieve the desired target poses. The complete execution process and detailed application scenarios are provided in the \href{\SuppVideoSII}{Supplementary Video S2} and \href{\SuppVideoSIII}{Supplementary Video S3}, respectively.

However, the current deployment of the \method exhibits limitations in initialization, which is critical for optimization convergence in highly non-convex problems. Even with advanced convexification techniques, initialization remains an open question in the broader optimization field~\cite{shetty2024tensor}. In this study, we demonstrate the effectiveness of the \method using random initialization, achieving notable success even with a 7-\ac{dof} manipulator and 64 target end-effector poses. However, task complexity increases the number of required initialization retries, as shown in \cref{tab:number_of_retries}. This escalation arises from the growing number of optimization variables, which impacts performance in two ways: it prolongs the time required to solve individual optimization instances and increases the total solution time due to additional retries, as illustrated in \cref{fig:run_time}. In more severe scenarios, like highly constrained environments, the random initialization could lead to failure.

Given these challenges, developing a more sophisticated initialization method for \method emerges as a promising direction for future work. Moreover, while this study focuses primarily on reachability and collision constraints, future work will explore incorporating additional real-world constraints, such as forces exerted at the end-effector, mounting limitations, and dynamic obstacles in the environment. These extensions will enhance \method's applicability to real-world scenarios.

\section{Conclusion}\label{sec:conclusion}

In this paper, we present \method, a novel optimization framework that improves both computational efficiency and solution optimality for base placement determination in fixed-base manipulators, outperforming traditional sampling-based methods. \method employs a two-layer optimization strategy: an outer layer for relaxed base exploration and an inner layer that addresses non-convex constraints using iterative local linearization. This approach effectively tackles challenges associated with long-horizon tasks, task-specific requirements, and self-collision avoidance. Furthermore, by operating directly in the configuration space, \method eliminates reliance on precomputed kinematic databases, providing a scalable and versatile solution for a wide range of robotic applications.

\noindent\textbf{Acknowledgment:}
We thank Dr. Chi Zhang (BIGAI), Yuyang Li (PKU), and Youqian Peng (PKU) for useful discussions.

\balance
\bibliography{reference_header_shorter,reference}
\bibliographystyle{IEEEtran}

\end{document}